\pdfoutput=1

\documentclass[11pt]{article}

\usepackage[]{coling}

\usepackage{times}
\usepackage{latexsym}

\usepackage[T1]{fontenc}

\usepackage[utf8]{inputenc}

\usepackage{microtype}

\usepackage{inconsolata}

\usepackage{graphicx}

\usepackage{url}
\usepackage{booktabs}
\usepackage{graphicx}
\usepackage{subcaption}
\usepackage{CJKutf8}
\usepackage{multirow}
\usepackage{xspace}
\usepackage{array}
\usepackage{amsmath} 
\usepackage{color, soul} 
\usepackage{algorithm, algorithmic}
\usepackage{enumitem}
\usepackage{xcolor}

\newcommand*{\affmark}[1][*]{\textsuperscript{\rm #1}}

\newcommand{\llama}{Llama-3-8B\xspace}
\newcommand{\mistral}{Mistral-7B\xspace}

\newcommand*{\affaddr}[1]{#1} 
\newcommand*{\email}[1]{\texttt{#1}}

\title{Zero-to-Strong Generalization: Eliciting Strong Capabilities of Large Language Models Iteratively without Gold Labels}

\author{
    Chaoqun Liu\thanks{Equal contribution. Chaoqun Liu and Qin Chao are under the Joint PhD Program between DAMO Academy and Nanyang Technological University. }\affmark[12]\;
    Qin Chao\affmark[*12]\;
    Wenxuan Zhang\thanks{Wenxuan Zhang is the corresponding author.}\affmark[23]\; 
    Xiaobao Wu\affmark[1]\; \\
    \textbf{Boyang Li}\affmark[1]\;
    \textbf{Anh Tuan Luu}\affmark[1]\;
    \textbf{Lidong Bing}\affmark[23]\\
    \affaddr{\affmark[1]Nanyang Technological University, Singapore};
    \affaddr{\affmark[2]DAMO Academy, Alibaba Group, Singapore}\\
    \affaddr{\affmark[3]Hupan Lab, 310023, Hangzhou, China}; \\
    \email{\{chaoqun.liu,qin.chao,saike.zwx,l.bing\}@alibaba-inc.com}\\
    \email{\{xiaobao002,boyang.li,anhtuan.luu\}@ntu.edu.sg}
    }

\begin{document}
\maketitle

\begin{abstract}
Large Language Models (LLMs) have demonstrated remarkable performance through supervised fine-tuning or in-context learning using gold labels. However, this paradigm is limited by the availability of gold labels, while in certain scenarios,
LLMs may need to perform tasks that are too complex for humans to provide such labels.
To tackle this challenge, this study explores whether solely utilizing unlabeled data can elicit strong model capabilities.
We propose a new paradigm termed \textit{zero-to-strong generalization}. We iteratively prompt LLMs to annotate unlabeled data and retain high-quality labels by filtering.
Surprisingly, we obverse that this iterative process gradually unlocks LLMs' potential on downstream tasks.
Our experiments on extensive classification and reasoning tasks confirm the effectiveness of our proposed framework.
Our analysis indicates that this paradigm is effective for both in-context learning and fine-tuning, and for various model sizes.

\end{abstract}
\section{Introduction}

Pre-trained language models (PLMs) have achieved significant improvements through supervised fine-tuning \cite{radford_improving_2018,devlin_bert_2019,wei_finetuned_2022,sanh_multitask_2022}. However, this paradigm often incurs high data costs and requires careful quality control. 
There are situations where advanced models need to tackle complex tasks that humans cannot fully comprehend or annotate. 
To study this problem, \citet{burns_weak--strong_2023}
consider the analogy of using weak models to supervise strong models. By fine-tuning the strong models on the labels generated by the weak supervisors, the strong student model consistently outperforms their weak supervisors, which they call \textit{weak-to-strong generalization}. This phenomenon occurs because strong pretrained models already possess good representations of relevant tasks.

Despite promising, this \textit{weak-to-strong generalization} paradigm has two limitations. Firstly, the student's performance is still constrained by the supervisor's ability to label data, and a weaker supervisor leads to a weaker student. 
Secondly, the reliance on weak supervisor models restricts its applicability to more scenarios. For example, there may be cases where no weak supervisors are available or humans cannot provide informative supervision in the future.

\begin{figure}[t]
    \centering
    \includegraphics[width=0.5\textwidth]{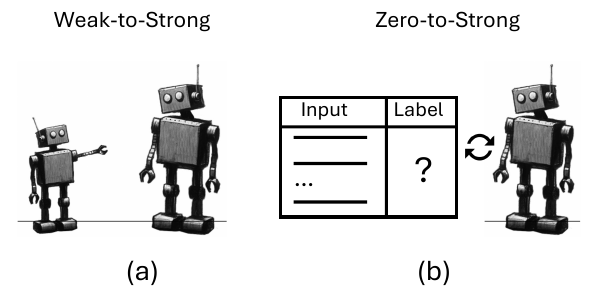}
    \caption{Illustration of (a) weak-to-strong \cite{burns_weak--strong_2023} and (b) our zero-to-strong analogy. While weak-to-strong uses weak models to supervise strong models, zero-to-strong elicits LLM capabilities without ground-truth labels or weak supervisors.}
    \label{fig:overview}
\end{figure}

\begin{figure*}[!ht]
    \centering
    \includegraphics[width=0.95\textwidth]{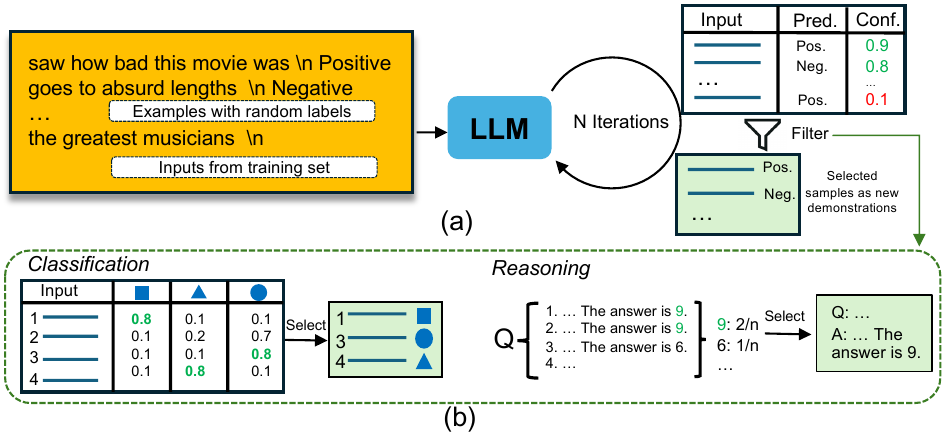}
    \caption{Illustration of (a) zero-to-strong generalization on a sentiment analysis task and (b) the filtering process. For classification tasks, we select demonstrations by ranking the probabilities for each label. For reasoning tasks, we select the most confident answers based on self-consistency \cite{wang_self-consistency_2023}.}
    \label{fig:framework}
\end{figure*}

To address the aforementioned issue, we explore how to harness the capabilities of LLMs without gold (or ground-truth) labels or weak supervisors, a process we refer to as \textit{zero-to-strong generalization}, as illustrated in Figure \ref{fig:overview}. Previous works have demonstrated that random labels \cite{min_rethinking_2022,yoo_ground-truth_2022} or invalid reasoning paths \cite{wang_towards_2023} can also yield good performance, although not as high as with gold labels. 
Inspired by this, we initially prompt LLMs with random or invalid demonstrations to label the data. We then select a new set of demonstrations based on confidence levels and prompt the LLMs again, repeating this process iteratively. This process allows us to achieve strong performance on tasks without needing gold-labeled data or weak supervisors.

We conducted experiments on 17 classification tasks, 2 extreme-label classification tasks, and 2 reasoning tasks to demonstrate the effectiveness of our proposed methods. Surprisingly, our method not only achieves performance comparable to but even outperforms in-context learning with gold labels for some tasks. 
We hypothesize that our method selects more suitable samples for demonstrations over iterations, which leads to high performance. 
Through careful analysis, we find that zero-to-strong learning is more effective for stronger models and more complex tasks. Additionally, it also works for fine-tuning and with larger models.

Our main contributions are summarized below: 
\begin{itemize}[leftmargin=*,itemsep=0pt]
    \item We propose a simple yet effective framework called zero-to-strong generalization, which elicits the strong capabilities of LLMs iteratively without gold labels. 
    \item
        We demonstrate the effectiveness of our zero-to-strong learning with extensive experiments on 17 classification tasks, 2 extreme-label classification tasks, and 2 reasoning tasks.
    \item
        We analyze the underlying reasons why zero-to-strong learning is effective and discover that its benefits extend to fine-tuning and larger models.
\end{itemize}

\section{Methodology}
This section begins with the problem definition, followed by our proposed zero-to-strong learning framework.

\subsection{Problem Definition}
In our setting, we assume the absence of gold labels, simulating situations where problems are so complex that human annotations are unreliable. However, we still possess minimal information about the problems. For instance, we know the label space  \(\mathcal{C}\) in a classification problem, and for a generation problem, the output format is defined. Additionally, we have access to a few inputs \(x_1, \ldots, x_k\) without gold labels.

\subsection{Zero-to-Strong Generalization} \label{sec:methods}

Figure \ref{fig:framework} illustrates our overall framework, comprising demonstration construction, response generation, sample selection, and iterative evolution.

\paragraph{Demonstration construction.} 

While we lack access to gold labels, we can create demonstrations by randomly sampling from the label space. For classification tasks, labels can be drawn as $\Tilde{y} \sim \mathcal{C}$. For reasoning tasks, we can manually generate outputs for a few examples, focusing on maintaining the correct format rather than ensuring complete accuracy. 

\paragraph{Response generation.}
The generated demonstrations are prepended to the input in the training set to form the LLM prompts. By prompting the LLMs, we generate both pseudo labels and their confidence for the training set samples. For classification tasks, we set the temperature to 0 and predict the labels using $\arg\max_{y \in \mathcal{C}} P(y|x)$, where $x$ is the text input and $\mathcal{C}$ is a limited set of potential labels. We use the normalized probability $P(y|x)$ as the confidence. For reasoning tasks, we set the temperature to 0.7 to sample diverse reasoning paths, selecting the most consistent final answer as the prediction. This method is similar to self-consistency \cite{wang_self-consistency_2023}, 
and we further calculate the ratio of consistent paths to the total number of paths as the confidence for each sample.

\paragraph{Sample selection.}
After generating the responses for all the training samples, we select the $k$ most confident samples for the next iteration. 
For classification tasks, we uniformly select the top-$k$ most confident samples across the label space. For reasoning tasks, we first identify the top-$k$ questions with the highest confidence.
Then, for each question, we randomly select one path from the consistent paths. The selection process is illustrated in Figure \ref{fig:framework}(b).

\paragraph{Iterative evolution.}
The selected samples and their predictions will serve as demonstrations for the next round, with this process repeating for several iterations and aiming for progressive performance improvement.

During the evaluation, we set the temperature to 0 and generate final predictions using the same method as in the response generation stage. The zero-to-strong algorithm for classification tasks is detailed in Algorithm \ref{algo:z2s} in Appendix \ref{app_sec:algo}.

\begin{table*}[ht]
\centering
\small
\resizebox{\textwidth}{!}{
\begin{tabular}{llccccccc}
\toprule
\multirow{2}{*}{Task} & \multirow{2}{*}{Setting} & \multicolumn{3}{c}{\llama} & & \multicolumn{3}{c}{\mistral} \\ \cmidrule{3-5} \cmidrule{7-9} 
      &      & 4-shot & 8-shot & 16-shot  && 4-shot & 8-shot & 16-shot \\ 
\midrule
\multirow{4}{*}{Classification} 
  & zero-shot                & 40.7   & 40.7   & 40.7    && 36.1   & 36.1   & 36.1    \\ 
  & random label             & 42.7   & 50.3   & 43.8    && 45.3   & 51.0   & 45.9    \\ 
  & gold label               & 53.3   & 56.6   & 61.1    && 57.5   & 57.5   & \textbf{60.5}    \\ \cmidrule{3-5} \cmidrule{7-9} 
  & ours (zero-to-strong)    & \textbf{57.5}   & \textbf{63.2}   & \textbf{61.4}    & &\textbf{61.1}   & \textbf{62.4}   & 60.1    \\ 

\midrule
\multirow{4}{*}{Extreme-label Classification} 
  & zero-shot                & 21.4   & 21.4   & 21.4    && \textbf{23.9}   & 23.9   & 23.9    \\ 
  & random label             & 4.5    & 3.7    & 2.5     && 5.3    & 3.6    & 2.3     \\ 
  & gold label               & 21.0   & 26.5   & 29.1    && 17.1   & \textbf{26.1}   & 26.4    \\ \cmidrule{3-5} \cmidrule{7-9} 
  & ours (zero-to-strong)    & \textbf{24.6}   & \textbf{27.2}   & \textbf{33.4}    & & 21.1 & 23.3   & \textbf{32.7}    \\ 

\bottomrule
\end{tabular}
}
\caption{Average Macro-F1 (\%) of \llama and \mistral on 17 classification and 2 extreme-label classification tasks.}
\label{tab:main_cls}
\end{table*}

\begin{table}[h]
\centering
\resizebox{0.48\textwidth}{!}{
\begin{tabular}{lcc}
\toprule
Setting                 & \llama & \mistral \\ 
\midrule
zero-shot               & 53.5       & 40.3       \\ 
invalid                 & 38.9       & 35.4       \\ 
gold label              & 62.2       & \textbf{53.4}       \\ 
ours (zero-to-strong)   & \textbf{64.2}       & 49.0 
\\
\bottomrule
\end{tabular}
}
\caption{Average accuracy (\%) of \llama and \mistral  on reasoning tasks.}
\label{tab:main_cot}
\end{table}

\section{Experiments}
We evaluate our proposed framework with two pretrained LLMs: Meta-Llama-3-8B (Llama-3-8B) \cite{dubey_llama_2024} and Mistral-7B-v0.1 (Mistral-7B) \cite{jiang_mistral_2023}. All the experiments are conducted on Nvdia A800 GPUs.

\subsection{Tasks}

We assess our framework's effectiveness through three tasks: standard text classification, extreme-label classification, and reasoning. Despite being a subtype of classification, extreme-label classification is treated separately due to its significantly larger class count.

\paragraph{Classification tasks.} 

Following \citet{yoo_ground-truth_2022}, we evaluate 17 widely-used text classification tasks, with dataset details in Table \ref{tab:stats_cls_elc} in the Appendix. Evaluations are conducted in 4-shot, 8-shot, and 16-shot, using manual templates from \citet{yoo_ground-truth_2022}.

\paragraph{Extreme-label classification tasks.}

Extreme-label classification poses greater challenges than traditional classification due to the large number of labels \cite{li_long-context_2024}. For evaluation, we selected the GoEmotions dataset with 28 classes \cite{demszky2020goemotions} and banking77 with 77 classes \cite{casanueva_efficient_2020}. Due to resource limitations, we sampled 1,000 instances from the training set and 500 from the test set. Dataset details can be found in Table \ref{tab:stats_extr_elc} in the Appendix.

\paragraph{Reasoning tasks.} 

We choose GSM8k \cite{cobbe2021gsm8k} and SVAMP \cite{patel_etal_2021_nlp} for evaluation, as both require multi-step reasoning. Details of the datasets are in Table \ref{tab:stats_reason_elc} in the Appendix. We selected up to 1,000 samples from the training set and used the entire test set for our experiment. Additionally, we generated 10 diverse reasoning paths for each sample during response generation.

\subsection{Baseline Methods}
We compare zero-to-strong with the following baseline methods: 
\paragraph{Zero-shot methods.} 

This setting does not use labeled data as demonstrations. For text and extreme-label classification tasks, predictions are made via $\arg\max_{y \in \mathcal{C}} P(y|x)$, where $x$ is the text input and $\mathcal{C}$ is a limited label set. For reasoning tasks, we adopt the Zero-shot-CoT approach \cite{kojima_large_2023}, prompting LLMs with "Let’s think step by step" and concluding with "Therefore, the answer (Arabic numerals) is" to obtain the final result.

\paragraph{Few-shot with gold labels.} 
For classification and extreme-label classification tasks, we sample $k$ input-label pairs $(x_1, y_1) \ldots (x_k, y_k)$ from the training set either randomly or uniformly based on the label space. We then make predictions via $\arg\max_{y \in \mathcal{C}} P(y|x_1, y_1 \ldots x_k, y_k, x)$. For reasoning tasks, we use a fixed set of demonstrations $(x_1, r_1, y_1) \ldots (x_k, r_k, y_k)$ to prompt LLMs, where $r_k$ represents the reasoning steps, following \citet{wei_chain--thought_2023}. The demonstrations are shown in Table \ref{tab:demo_cot_gold} in the Appendix. The final answer is extracted using a regular expression.

\paragraph{Few-shot with invalid labels.} In classification and extreme-label classification, demonstrations are generated by assigning random labels rather than using the actual data labels. Each $x_i$ ($1 \leq i \leq k$) is paired with a randomly sampled label $\tilde{y}_i$ from $\mathcal{C}$. The sequence $(x_1, \tilde{y}_1) \ldots (x_k, \tilde{y}_k)$ is then used to make a prediction by maximizing $\arg\max_{y \in \mathcal{C}} P(y|x_1, \tilde{y}_1 \ldots x_k, \tilde{y}_k, x)$. For reasoning tasks, we reused demonstrations with the "no coherence" setting \cite{wang_towards_2023}, meaning the rationales are out of order, as shown in Table \ref{tab:demo_cot_no_co} in the Appendix.

To ensure reproducibility, we set the evaluation temperature to 0. Results for gold-label, invalid labels, and zero-to-strong are averaged over three seeds to sample the training set for demonstrations. For methods other than zero-shot, initial demonstrations are sampled using two approaches: 1) random initialization --- random sampling from the training set, and 2) uniform initialization --- sampling an equal number of instances from each class.

\begin{figure*}[t]
    \centering
    \begin{subfigure}[b]{0.85\textwidth}
        \includegraphics[width=\textwidth]{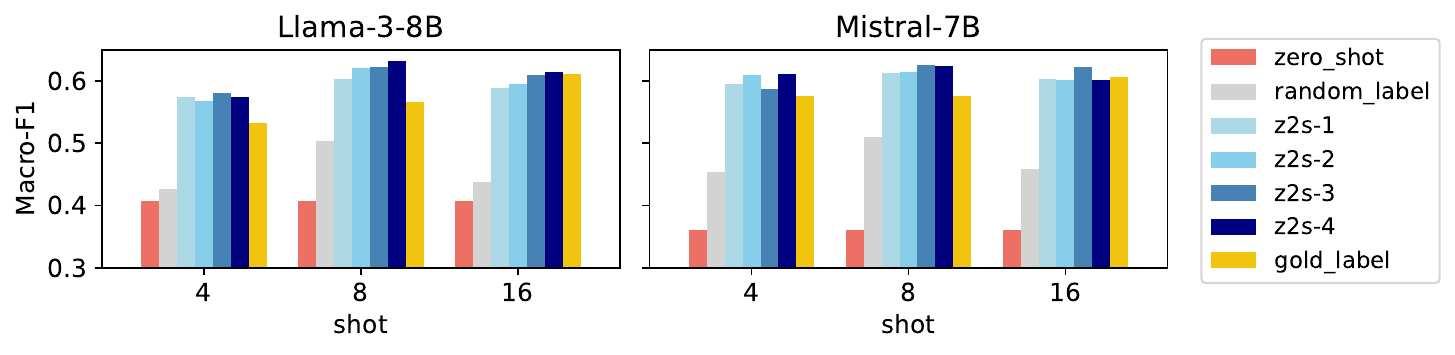}
        \caption{Random initialization.}
        \label{fig:sub1}
    \end{subfigure}
    \begin{subfigure}[b]{0.85\textwidth}
        \includegraphics[width=\textwidth]{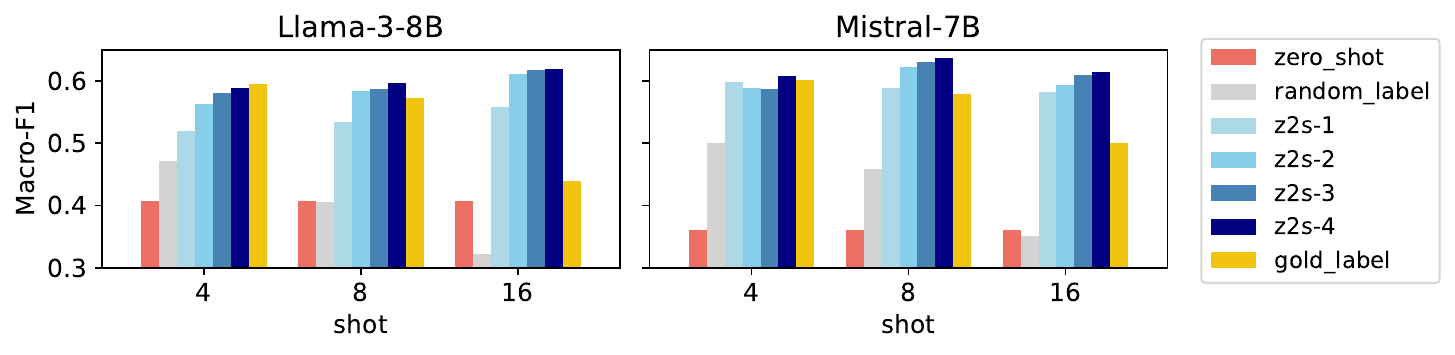}
        \caption{Uniform initialization.}
        \label{fig:sub1}
    \end{subfigure}
    \caption{Average macro-F1 for 17 classification tasks, using two LLMs and two initialization settings. "z2s-\textit{i}" means the \textit{i}th round of iteration for zero-to-strong method.}
    \label{fig:results_cls}
\end{figure*}

\begin{figure*}[t]
    \centering
    \begin{subfigure}[b]{0.85\textwidth}
        \includegraphics[width=\textwidth]{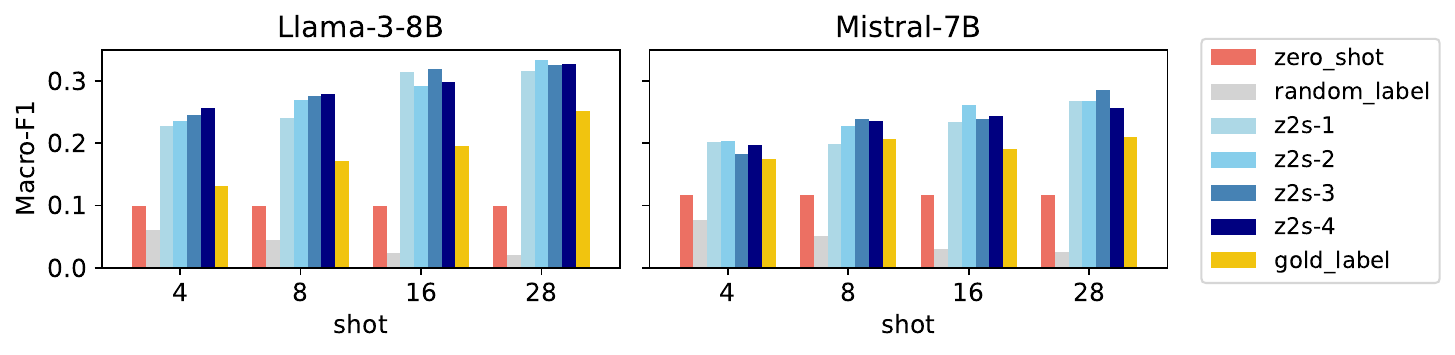}
        \caption{Random initialization.}
        \label{fig:sub1}
    \end{subfigure}
    \begin{subfigure}[b]{0.85\textwidth}
        \includegraphics[width=\textwidth]{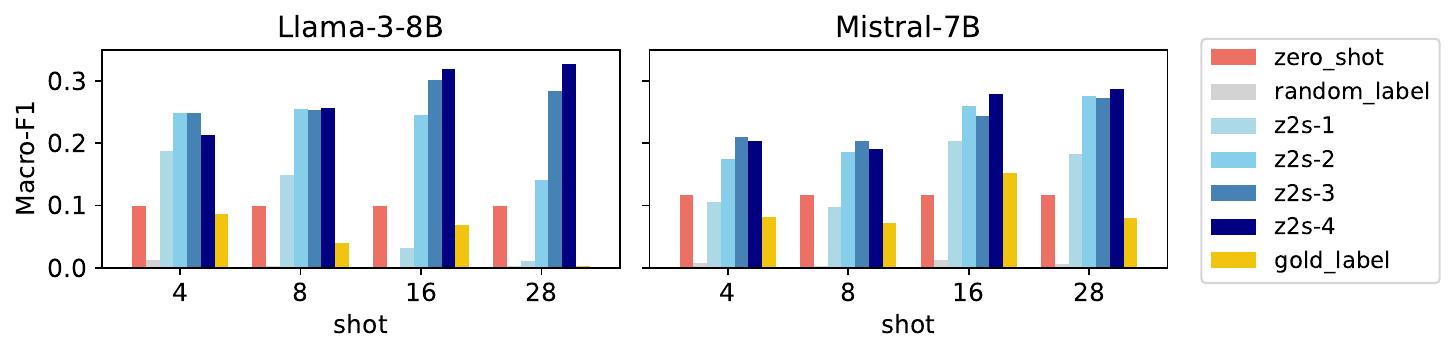}
        \caption{Uniform initialization.}
        \label{fig:sub1}
    \end{subfigure}
    \caption{Average macro-F1 for GoEmotions, using two LLMs and two initialization settings.}
    \label{fig:results_go}
\end{figure*}

\subsection{Main Results}\label{sec:main_results}

Table \ref{tab:main_cls} presents the main results for classification and extreme-label classification tasks. Our zero-to-strong method for \llama consistently outperforms other approaches across all shots settings, demonstrating its effectiveness. It also yields the best results with shots lower than 16 for \mistral. We believe this difference stems from \llama's superior capabilities, as zero-to-strong performance relies on inherent capabilities gained during pre-training. Overall, extreme-label classification tasks show lower performance compared to standard tasks, emphasizing their increased difficulty. Poor performance in random label settings underlines the necessity of accurate labels for these challenging tasks. Additionally, the number of demonstrations significantly affects extreme-label classification, as performance with gold-label and zero-to-strong settings improves with more demonstrations, while random-label performance declines.

Table \ref{tab:main_cot} presents the average accuracies for the two reasoning tasks. Our zero-to-strong method outperforms other approaches using \llama, yet it still lags behind the few-shot method with gold labels using \mistral. This trend aligns with classification and extreme-label classification results, indicating that zero-to-strong is more effective with stronger models. As models continue to improve in the future, our approach may gain even more advantages.

\subsection{Analysis}
The zero-to-strong performance is promising. To better understand its behavior and underlying reasons, we conduct the following analysis.

\subsubsection{How does the performance improve over the iterations?}
\paragraph{Classification tasks.} The detailed results for 17 classification tasks are shown in Figure \ref{fig:results_cls}. It can be seen that for both models, zero-to-strong can achieve comparable or better results than few-shot with gold labels within 4 rounds of iteration. We hypothesize that the zero-to-strong method selects the most confident samples as demonstrations, which is superior to randomly sampling from gold labels. Zero-to-strong also has a big advantage over few-shot with random labels (please note that few-shot with random labels can be regarded as the 0th round for zero-to-strong). We also notice that for some settings LLMs improve iteration by iteration but the benefits diminish after certain rounds and the performances fluctuate. In addition, the phenomenon exists for all numbers of shots.

\paragraph{Extreme label classification.}
The results for GoEmotions are shown in Figure \ref{fig:results_go} and the results for banking77 are shown in Figure  \ref{fig:results_banking} in the Appendix.
With more demonstrations, few-shots with gold labels perform better with random initialization. It is interesting that when the number of shots is small, few-shot with gold labels underperforms zero-shot setting. We hypothesize that when the number of shots is small, it cannot cover all the labels and make the distribution of the demonstration deviate from the test set. For few-shot with random labels, more demonstrations hurt the performance. This is reasonable as more demonstrations result in more wrong demonstrations, which deteriorate performance. Interestingly, zero-to-strong outperforms few-shot with gold labels in all settings for GoEmotions but the relative performance depends on the initialization settings and the number of shots, which again confirms the effectiveness of zero-to-strong method.

\begin{figure}[]
    \centering
    \includegraphics[width=0.95\linewidth]{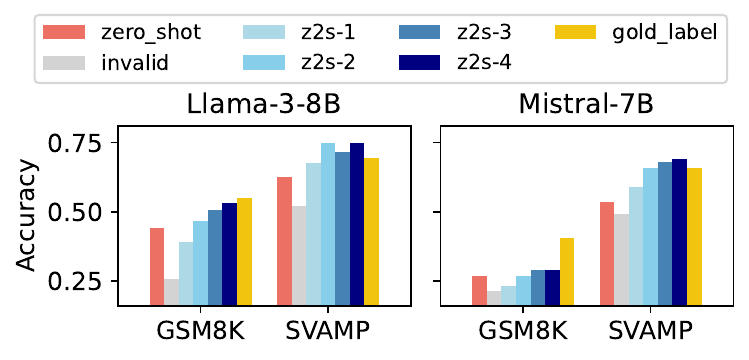}
    \caption{Accuracy for the two reasoning tasks.}
    \label{fig:results_cot}
\end{figure}

\paragraph{Reasoning tasks.}
The results for the two reasoning tasks are shown in Figure \ref{fig:results_cot}. For GSM8K, zero-to-strong improves performance iteration by iteration and approaches few-shot with gold labels after 4 iterations. For SVAMP, zero-to-strong outperforms few-shot with gold labels after a few iterations. We hypothesize that the initial demonstrations with gold label are not optimal for SVAMP and we can generate better demonstrations for this task with zero-to-strong approach.

\subsubsection{What happens during the iterations?}
To further understand the mechanics behind zero-to-strong approach, we conduct more analysis on GoEmotions and GSM8K.

\begin{figure*}[t]
    \centering
    \begin{subfigure}[b]{\textwidth}
        \includegraphics[width=0.245\textwidth]{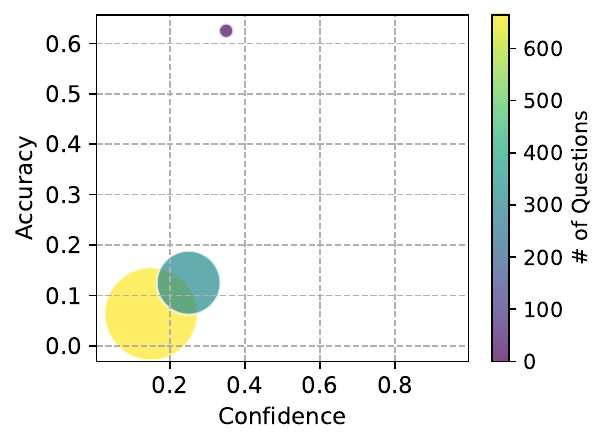}
        \includegraphics[width=0.245\textwidth]{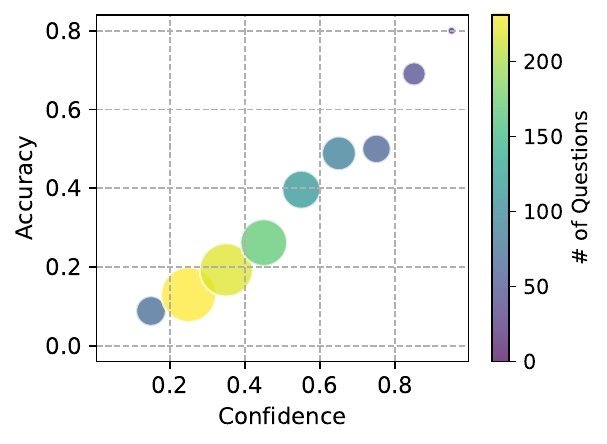}
        \includegraphics[width=0.245\textwidth]{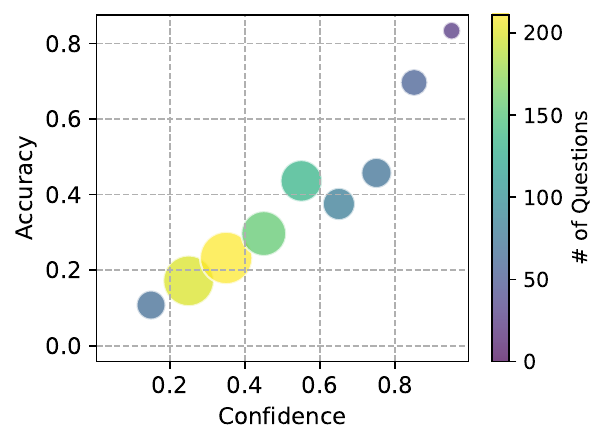}
        \includegraphics[width=0.245\textwidth]{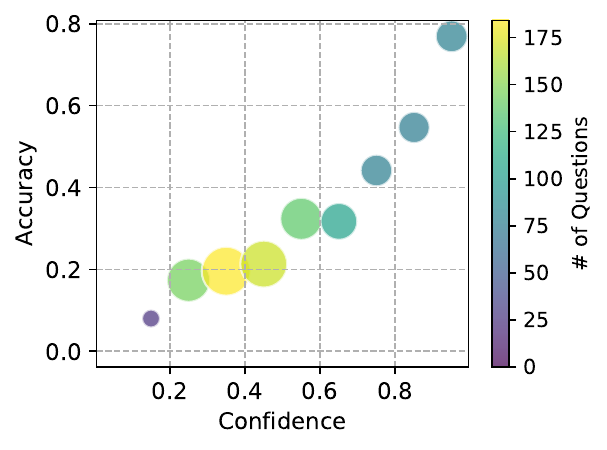}
        \caption{Accuracy vs confidence for GoEmotions from iteration 1 (left) to iteration 4 (right).}
        \label{fig:sub1}
    \end{subfigure}
    \begin{subfigure}[b]{\textwidth}
        \includegraphics[width=0.245\textwidth]{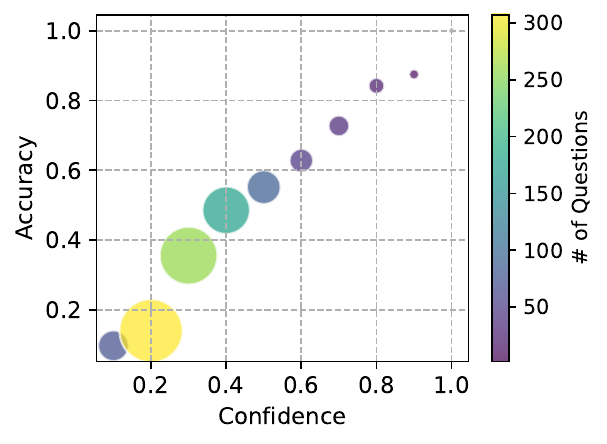}
        \includegraphics[width=0.245\textwidth]{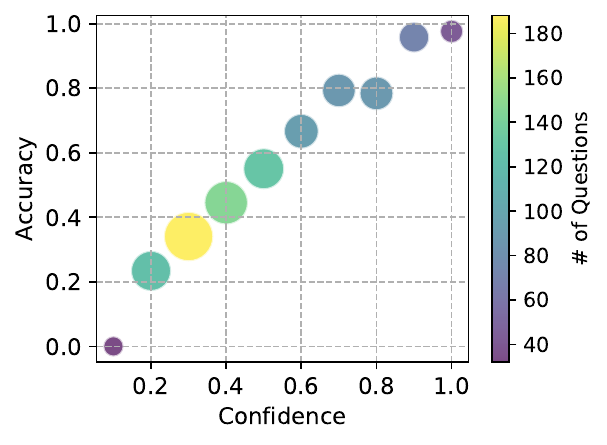}
        \includegraphics[width=0.245\textwidth]{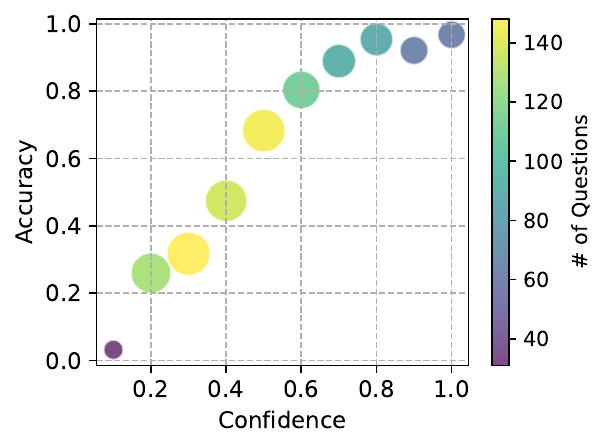}
        \includegraphics[width=0.245\textwidth]{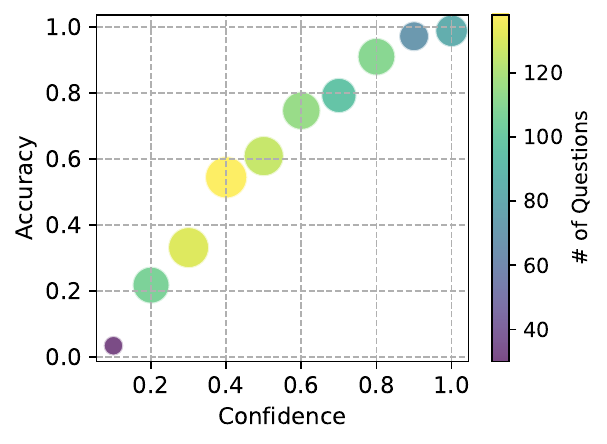}
        \caption{Accuracy vs confidence for GSM8K from iteration 1 (left) to iteration 4 (right).}
        \label{fig:sub2}
    \end{subfigure}
    \caption{The relation between accuracy and confidence of the answers for the training set from iteration 1 to iteration 4. The confidence of GoEmotions and GSM8K is calculated based on the methods described in Section \ref{sec:methods}. After each iteration, more samples are becoming more confident and accurate.}
    \label{fig:analysis_confidence}
\end{figure*}

\paragraph{Does the confidence correlate with the accuracy?}
Our sample selection process is based on the hypothesis that predictions with higher confidence will have higher accuracy. To verify this hypothesis, we plot the distributions of the sample confidence and their accuracy in Figure \ref{fig:analysis_confidence}. It can be seen that accuracy is highly correlated with confidence. Initially, more samples have low confidence and low accuracy. After several iterations, more samples have higher confidence and higher accuracy. This observation explains why the model performs better and better. 

\begin{figure}[ht]
    \centering
    \includegraphics[width=0.7\linewidth]{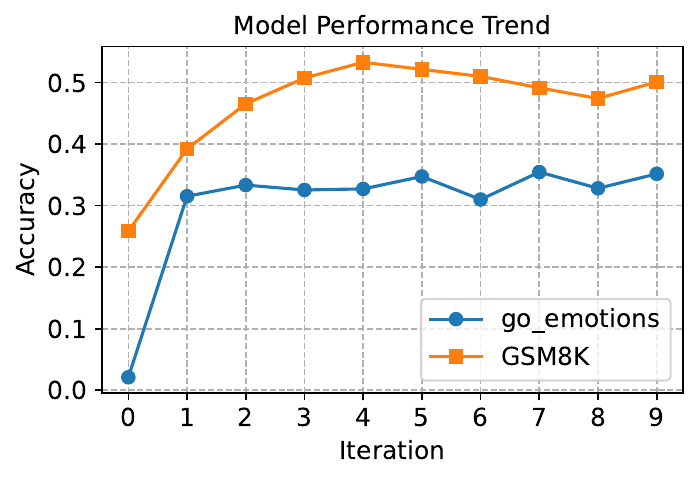}
    \caption{The accuracy for more iterations for zero-to-strong on GSM8K and GoEmotions. The evaluation is on Llama-3-8B.}
    \label{fig:more_iterations}
\end{figure}

\paragraph{Do more iterations help with the final performance?}
In Section \ref{sec:main_results}, we initially set the maximum number of iterations to 4. In some cases, performance consistently improved with each iteration. However, in other cases, performance reached a plateau after a certain number of iterations and subsequently fluctuated. To further explore the models' performance over a greater number of iterations, we extended the total number of iterations to 9. The results, depicted in Figure \ref{fig:more_iterations}, indicate that performance does not improve beyond a certain point. We hypothesize that once the optimal demonstrations are selected, additional iterations do not contribute to further improvements.

\paragraph{Are the demonstrations more and more confident and accurate over iterations?}
We select the demonstrations for the next iteration based on confidence. Thus we expect the confidence to increase over iterations. As shown in Figure \ref{fig:ana_demo_go_random}, \ref{fig:ana_demo_go_uniform} and \ref{fig:ana_demo_gsm} (in the Appendix), the confidence for both GoEmotions and GSM8K increases steadily but saturates after a few iterations. For GoEmotions, confidence for the smaller number of shots is larger and saturates faster. This is expected, as it is harder to get more confident samples. It is also interesting that for GoEmotions, random initialization converges faster than uniform initialization, which is also observed in Figure \ref{fig:results_go}. The possible reason is that the training set is not uniform, thus it is better to initialize the demonstrations randomly.

Even though we select the most confident samples for each iteration, we cannot guarantee the accuracy of the selected demonstrations. As shown in Figure \ref{fig:ana_demo_go_random}(b) and \ref{fig:ana_demo_go_uniform}(b), the accuracy of the demonstrations fluctuates or even decreases after certain iterations. This is a possible reason why the performances on evaluation sets fluctuate after certain iterations.

\begin{figure}[t]
    \centering
    \begin{subfigure}[b]{0.23\textwidth}
        \includegraphics[width=\textwidth]{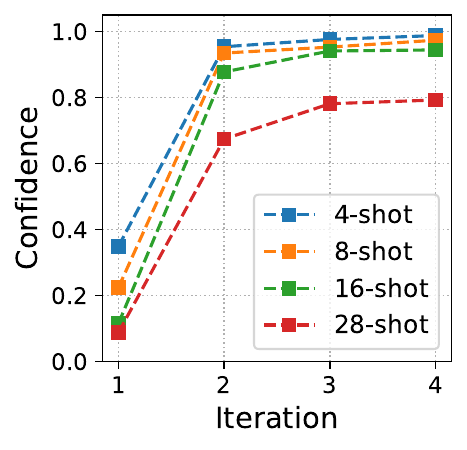}
        \caption{Confidence}
        \label{fig:sub1}
    \end{subfigure}
    \begin{subfigure}[b]{0.23\textwidth}
        \includegraphics[width=\textwidth]{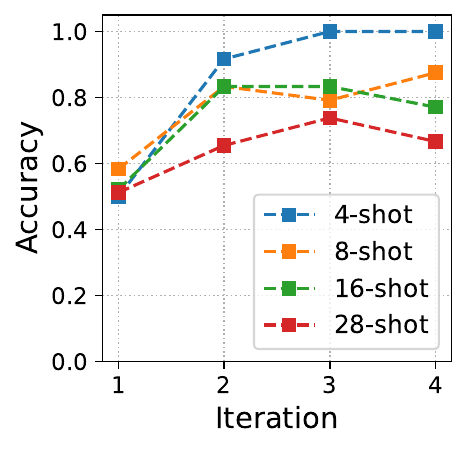}
        \caption{Accuracy}
        \label{fig:sub2}
    \end{subfigure}
    \caption{Confidence and accuracy of demonstrations over iterations for GoEmotions with random initialization.}
    \label{fig:ana_demo_go_random}
\end{figure}

\begin{figure}[t]
    \centering
    \begin{subfigure}[b]{0.23\textwidth}
        \includegraphics[width=\textwidth]{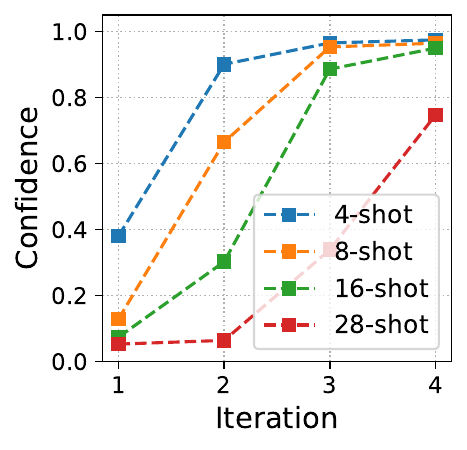}
        \caption{Confidence}
        \label{fig:sub1}
    \end{subfigure}
    \begin{subfigure}[b]{0.23\textwidth}
        \includegraphics[width=\textwidth]{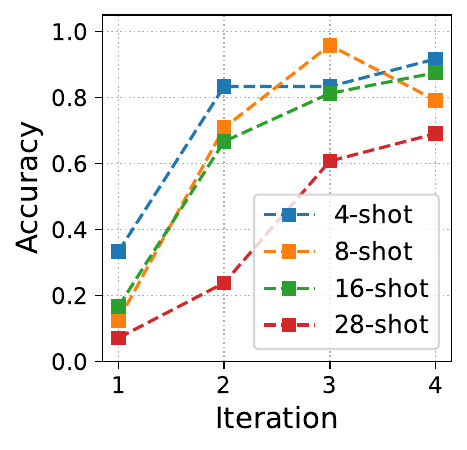}
        \caption{Accuracy}
        \label{fig:sub2}
    \end{subfigure}
    \caption{Confidence and accuracy of demonstrations over iterations for GoEmotions with uniform initialization.}
    \label{fig:ana_demo_go_uniform}
\end{figure}

\begin{table}[!t]
    \centering
    \small
    \setlength{\tabcolsep}{3pt}
    \begin{tabular}{lcccccc}
    \toprule   
        Setting & invalid & z2s-1 & z2s-2 & z2s-3 & z2s-4 \\ 
        \midrule
        Invalid Reasoning & 48.8 & 51.6 & 54.5 & 50.7 & 51.9 \\ 
        No coherence & 25.9 & 46.7 & 51.0 & 46.6 & 51.8 \\ 
        No coherence for BOs & 43.9 & 52.6 & 54.7 & 54.2 & 53.8 \\ 
        No coherence for LTs & 29.0 & 47.3 & 52.8 & 52.7 & 50.3 \\ 
        No relevance & 3.9 & 2.7 & 2.6 & 2.7 & 2.8 \\ 
        No relevance for BOs & 37.0 & 53.2 & 49.6 & 51.9 & 51.6 \\ 
        No relevance for LTs & 27.8 & 47.7 & 49.4 & 49.5 & 51.9 \\ 
        \midrule
        Invalid RnA & 38.1 & 46.0 & 51.4 & 51.0 & 50.9 \\ 
    \bottomrule
    \end{tabular}
    \caption{GSM8K with different invalid demonstrations for Llama-3-8B. The zero\_shot score is 44.3, while the few-shot with gold\_label is 55.0. "BO" refers to bridging objects and "LT" refers to "language templates". "RnA" refers to "reasoning and answer".}
    \label{tab:ana_cot_demos}
\end{table}


\paragraph{Does it work with different initial demonstrations for reasoning tasks?}
In the previous experiments, we used the "no coherence" demonstration for initialization. To evaluate whether our method applies to general incorrect demonstrations, we tested other settings from \citet{wang-etal-2022-towards}. Additionally, we manually created a new set of demonstrations featuring invalid reasoning and incorrect final answers but containing relevant bridging objects and language templates, as illustrated in Table \ref{tab:demo_cot_manual} in the Appendix. We generate 5 reasoning paths during response generation for this analysis.
The results are presented in Table \ref{tab:ana_cot_demos}. From the results, it is evident that the zero-to-strong method achieves accuracies greater than 50\% across all settings, except for the "no relevance" condition. This indicates that providing relevant demonstrations is crucial for the zero-to-strong method to be effective. Fortunately, this requirement is manageable for humans, as providing incorrect but relevant reasoning paths and final answers is not hard.

\subsubsection{Does it work for fine-tuning besides in-context learning?}\label{sec:extend_2_finetune}

We further investigate the impact of incorporating fine-tuning with LoRA \cite{hu_lora_2021} into our framework. We first generate the labels for the training set with ICL and demonstrations with random labels. Then we filter the samples and fine-tune the model with the pseudo training set. After that, we generate the new labels with the fine-tuned model in a zero-shot manner. We repeat the above process for several iterations, as detailed in Appendix \ref{appen:fine-tuning}. Optionally, we can fine-tune the model with samples labeled after four rounds of zero-to-strong with ICL.
As shown in Table \ref{tab:finetune_4iter}, fine-tuning also improves progressively, notably surpassing few-shot results with gold labels for GoEmotions.

\begin{table}[!t]
    \centering
    \small
    \setlength{\tabcolsep}{3.5pt}
    \begin{tabular}{lccccccc}
    \toprule   
tasks & ZS & ft1 & ft2 & ft3 & ft4  & z2s-4+ft &GL \\
\midrule
GoE & 9.9 &25.3 & 26.6 & 26.7 & 26.0  & 31.7 & 17.2 \\
GSM8K & 44.3 & 30.2 & 49.7 & 51.1 & 50.3 & 50.3 & 55.9\\
    \bottomrule
    \end{tabular}
    \caption{Fine-tuning performance for Llama-3-8B. "GoE" refers to "GoEmotions". 
    Results are averaged over 3 seeds. "ZS" refers to "zero-shot". "ft" stands for "fine-tuning".   "GL" refers to "gold label".}
    \label{tab:finetune_4iter}
\end{table}


\subsubsection{Does it work for larger models?}\label{sec:extend_2_larger}
Even though smaller LLMs are more computationally efficient, larger models normally have better performances. 
To assess the effectiveness of our approach on larger models, we evaluated it on two larger models: Meta-Llama-3-70B (Llama-3-70B) \cite{dubey_llama_2024} and Mixtral-8x22B-v0.1 (Mixtral-8x22B) \cite{jiang_mixtral_2024} on GSM8K. As shown in Table \ref{tab:large_models}, zero-to-strong with the two models outperforms the zero-shot setting and achieves comparable performance with few-shot with gold labels, which is consistent with that observed on smaller models, suggesting that our method generalizes well across models of varying sizes.

\begin{table}[!t]
    \centering
    \small
    \setlength{\tabcolsep}{2pt}
    \begin{tabular}{lccccccc}
    \toprule   
        model & ZS & INV & z2s-1 & z2s-2 & z2s-3 & z2s-4 & GL \\ 
        \midrule
        Llama-3-70B & 73.7 & 30.3 & 60.7 & 76.7 & 80.1 & 80.7 & 82.3 \\ 
        Mixtral-8x22B & 61.0 & 19.3 & 56.7 & 71.2 & 72.4 & 69.8 & 67.9 \\ 
    \bottomrule
    \end{tabular}
    \caption{Accuracies on GSM8K with larger models. "ZS" refers to "zero-shot". "INV" refers to "invalid". "GL" refers to "gold label".}
    \label{tab:large_models}
\end{table}

\section{Related Work}
\paragraph{Weak-to-strong generalization.}
In the future, advanced models will handle complex tasks with only weak human supervision. To study this, \citet{burns_weak--strong_2023} proposed using weak supervisor models to elicit the capabilities of stronger student models. Their findings revealed that, after fine-tuning, the strong student models consistently outperformed the weak supervisor models, a phenomenon they term \textit{weak-to-strong generalization}. In contrast to transferring knowledge from strong models to models \cite{meng_generating_2022,ye_zerogen_2022},
this learning paradigm is a specific type of weakly-supervised learning \cite{bach2017learning}, where models are trained with noisy or biased labels \cite{bellamy2019ai, song2022learning, liu-etal-2023-zero}. Our work eliminates the necessity of weak models or weak labels for supervision. Instead, we utilize minimal supervision, such as the label space or incorrect initial demonstrations, to elicit the capabilities of large language models. Other research has proposed self-improvement of LLMs using labeled or unlabeled data \cite{huang2022large,li2023mot,zelikman_star_2022} for reasoning tasks. In contrast, we aim to propose a general framework for learning new tasks without labeled data.

\paragraph{Understanding In-context learning.}
In-context learning (ICL) \cite{brown_language_2020} can effectively learn new tasks with a few demonstrations, but its mechanism is still under discussion. Previous research \cite{lu2021fantastically,zhao2021calibrate,su2022selective} found that ICL is sensitive to the demonstration samples, their order, and their diversity. Studies by \citet{min_rethinking_2022} and \citet{wang_towards_2023} discovered that even random labels for classification or invalid demonstrations for reasoning tasks can yield good performance, suggesting that gold labels are not always necessary. However, \citet{yoo_ground-truth_2022} showed that correct input-label mappings can have varying impacts through extensive experiments. Recently, \citet{wang_learning_2024} found that learning to retrieve in-context examples helps improve the performance, but the gold labels are needed. In contrast, our work achieves strong performance with random or invalid labels and further improves iteratively to attain even better results.

\section{Conclusion}
In this work, we propose a new framework called \textit{zero-to-strong generalization}. Without gold label data or weaker supervisors, we can elicit the capabilities of LLMs iteratively through prompting and filtering. Experiments on classification and reasoning tasks demonstrate the effectiveness of this framework.
Further analysis shows that by selecting the most confident samples as demonstrations for the next iteration, we also select more accurate and more suitable demonstrations. This framework also generalizes well to fine-tuning and larger models. Our work demonstrates the feasibility of eliciting the capabilities of LLMs with minimal supervision. In the future, we plan to explore \textit{zero-to-strong generalization} in more diverse and challenging tasks.

\section*{Limitations}
Our framework is restricted to tasks with a single definitive correct answer. For instance, sentences in glue-sst2 \cite{socher2013recursive} can be either positive or negative, and the final answer in GSM8k \cite{cobbe_training_2021} must be a single number. This uniqueness of the final answer allows us to calculate the confidence of the generated responses. However, for open-ended tasks like story writing, our method is not applicable, as we cannot determine the confidence level of the generated content and leave this as future work. 

\section*{Acknowledgements}
This research is supported, in part, by DAMO Academy through DAMO Academy Research Intern Program and Alibaba-NTU Singapore Joint Research Institute (JRI), Nanyang Technological University, Singapore. Chaoqun Liu extends his gratitude to Interdisciplinary Graduate Programme and College of Computing and Data Science, Nanyang Technological University, Singapore, for their support. 

\bibliography{anthology,custom, references1}

\appendix
\section{Appendix}
\label{sec:appendix}

\subsection{Methodology} \label{app_sec:algo}
The algorithm for zero-to-strong on classification tasks is shown in Algorithm \ref{algo:z2s}.

\begin{algorithm}[]
\small
\caption{Zero-to-Strong}
\begin{algorithmic}[1]
\REQUIRE A $LLM$ with $\Pr(y|x)$ accessible.
\REQUIRE Input data $X$, and the label space $\mathcal{C}$
\REQUIRE Max iterations $M$, number of demos $K$
\STATE Initial state: $D_{0}$, contains $K$ random labelled demonstrations from $X$
\WHILE{Iter $t$ < $M$}
    \STATE Calculate $\hat{y} = \arg\max_{y \in \mathcal{C}} P(y|D_{t-1};x)$;
    \STATE Sort the $\hat{Y} = \{\hat{y_1},..., \hat{y_i}\}$ in descending order of probability;
    \STATE $D_t= \{\}$
    \WHILE{ $|D_t| < K$}
    \IF{$\hat{y_i} \not\in D_{t-i}$}
    \STATE $D_t  = D_t \cup \hat{y}$;
    \STATE $i=i+1$;
    \ENDIF
    \ENDWHILE    
\ENDWHILE
\RETURN $\hat{Y}$
\end{algorithmic}
\label{algo:z2s}
\end{algorithm}

\subsection{Experiment Setup}
The list and statistics of 17 classification tasks, 2 extreme-label classification tasks, and 2 reasoning tasks are shown in Table \ref{tab:stats_cls_elc}, \ref{tab:stats_extr_elc} and \ref{tab:stats_reason_elc}, respectively. The 17 text classification datasets span a variety of tasks such as sentiment analysis, paraphrase detection, natural language inference, and hate speech detection. GoEmotions is an emotion classification task and banking77 is an intent classification task.

\begin{table}[h]
    \centering
    \resizebox{0.99\columnwidth}{!}{%
        \begin{tabular}{lrrr}
        \toprule
        Dataset & \#Train & \#Test & \#C \\\midrule
        glue-sst2 \cite{socher2013recursive} & 67,349 & 872 & 2 \\
        glue-rte \cite{dagan2005pascal} & 2,490 & 277 & 2 \\
        glue-mrpc \cite{dolan2005automatically} & 3,668 & 408 & 2 \\
        glue-wnli \cite{levesque2012winograd} & 635 & 71 & 2 \\
        super\_glue-cb \cite{de2019commitmentbank} & 250 & 56 & 3 \\
        trec \cite{voorhees2000building} & 5,452 & 500 & 5 \\
        financial\_phrasebank \cite{malo2014good} & 1,181 & 453 & 3 \\
        poem\_sentiment \cite{sheng2020investigating} & 843 & 105 & 3 \\
        medical\_questions\_pairs \cite{mccreery2020effective} & 2,438 & 610 & 2 \\
        sick \cite{marelli2014sick} & 4,439 & 495 & 3 \\
        hate\_speech18 \cite{de2018hate} & 8,562 & 2,141 & 4 \\
        ethos-national\_origin \cite{mollas2020ethos} & 346 & 87 & 2 \\
        ethos-race \cite{mollas2020ethos} & 346 & 87 & 2 \\
        ethos-religion \cite{mollas2020ethos} & 346 & 87 & 2 \\
        tweet\_eval-hate \cite{barbieri2020tweeteval} & 9,000 & 1,000 & 2 \\
        tweet\_eval-stance\_atheism \cite{barbieri2020tweeteval} & 461 & 52 & 3 \\
        tweet\_eval-stance\_feminist \cite{barbieri2020tweeteval} & 597 & 67 & 3 \\ \bottomrule
        \end{tabular}
    }
    \caption{Data splits of the 17 classification tasks (\#C means number of classes.)}
    \label{tab:stats_cls_elc}
\end{table}

\begin{table}[h]
    \centering
    \small
    \resizebox{0.99\columnwidth}{!}{%
    \begin{tabular}{cccc}
        \toprule
    Dataset & \#Train & \#Test & \#Classes \\
    \midrule
    GoEmotions \cite{demszky2020goemotions} &  36308 &  4590 & 28 \\
    banking77  \cite{casanueva_efficient_2020}  & 10003 & 3080 & 77 \\
    \bottomrule
    \end{tabular}
    }
    \caption{Data splits of the 2 extreme-label classification tasks.}
    \label{tab:stats_extr_elc}
\end{table}

\begin{table}[h]
    \centering
    \small
    \begin{tabular}{ccc}
        \toprule
    Dataset & \# Train & \# Test \\
    \midrule
    GSM8k \cite{cobbe2021gsm8k}  & 7473 & 1319 \\
    	
    SVAMP \cite{patel_etal_2021_nlp} &  700 & 300 \\
    \bottomrule
    \end{tabular}
    \caption{Data splits of the 2 reasoning tasks.}
    \label{tab:stats_reason_elc}
\end{table}

\begin{figure}[t]
    \centering
    \begin{subfigure}[b]{0.23\textwidth}
        \includegraphics[width=\textwidth]{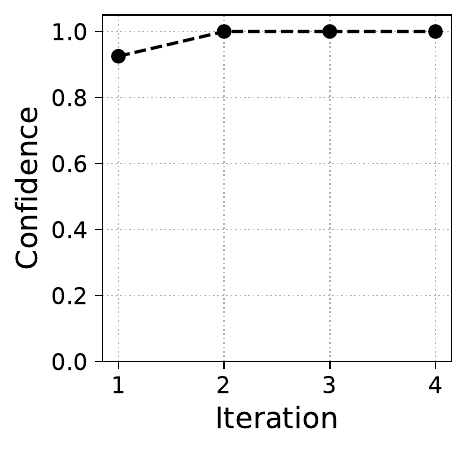}
        \caption{Confidence}
        \label{fig:sub1}
    \end{subfigure}
    \begin{subfigure}[b]{0.23\textwidth}
        \includegraphics[width=\textwidth]{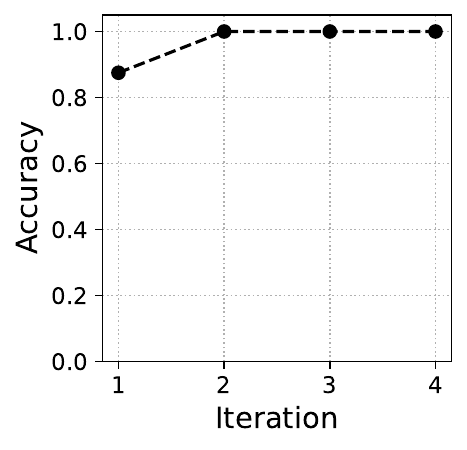}
        \caption{Accuracy}
        \label{fig:sub2}
    \end{subfigure}
    \caption{Confidence and accuracy of demonstrations over iterations for GSM8K.}
    \label{fig:ana_demo_gsm}
\end{figure}

\begin{figure*}[t]
    \centering
    \begin{subfigure}[b]{0.9\textwidth}
        \includegraphics[width=\textwidth]{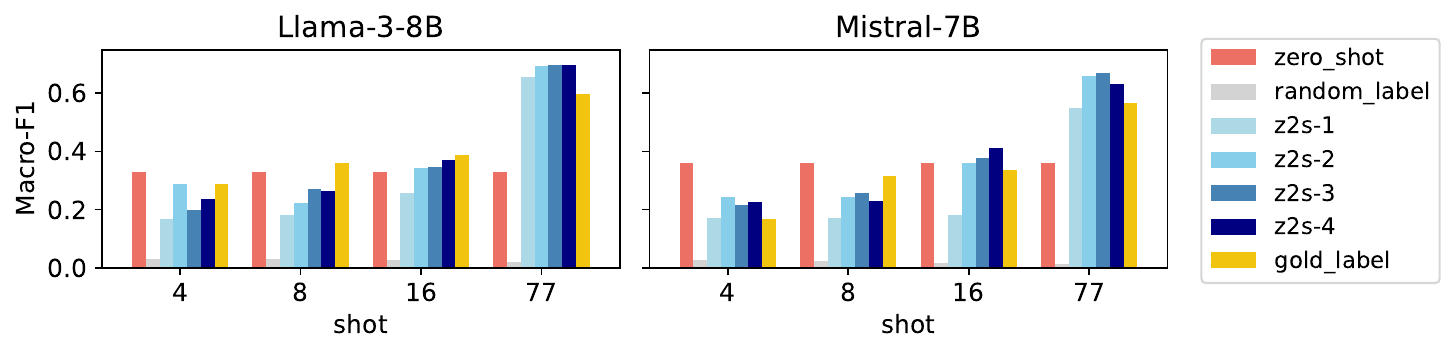}
        \caption{Random initialization.}
        \label{fig:sub1}
    \end{subfigure}
    \begin{subfigure}[b]{0.9\textwidth}
        \includegraphics[width=\textwidth]{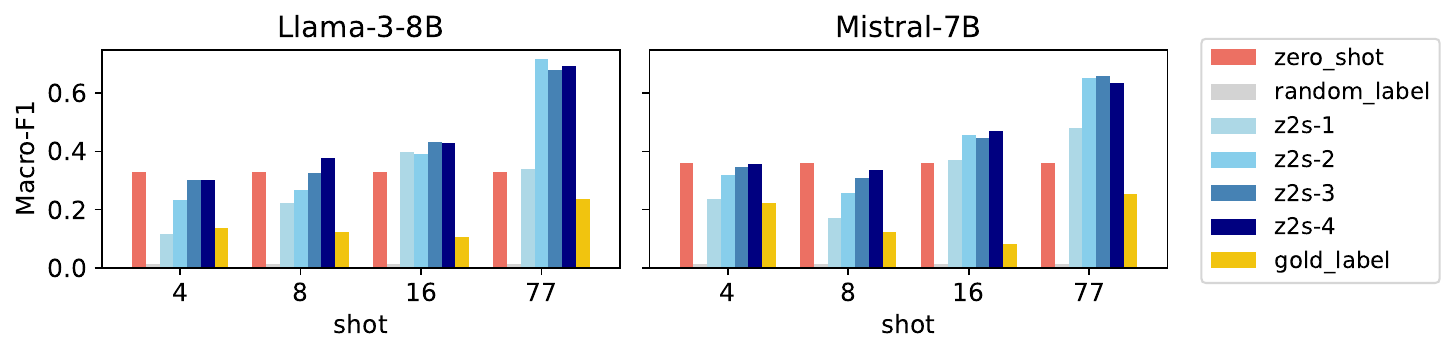}
        \caption{Uniform initialization.}
        \label{fig:sub1}
    \end{subfigure}
    \caption{Average macro-F1 for banking77, using two LLMs and two initialization settings. }
    \label{fig:results_banking}
\end{figure*}

For the 17 classification tasks, we adopt the manual templates and verbalizers from \citealp{yoo_ground-truth_2022} if possible. Examples for some tasks are shown in Table \ref{tab:template-cls}. The templates for the two extreme-classification tasks are shown in Table \ref{tab:template_elc}. The newly created template for "invalid reasoning and answer" is shown in Table \ref{tab:demo_cot_manual}. We keep all the questions the same and modify the reasoning paths and the final answer to make sure they are wrong. 

\subsubsection{Fine-tuning setup} \label{appen:fine-tuning}

For the fine-tuning experiments, as mentioned in Section \ref{sec:extend_2_finetune}, we filter out low-quality training data before each iteration of fine-tuning. For GoEmotions, according to the probability, we retain only the top $\frac{1}{|C|}$ data points for each class from the label space $C$. This results in duplicated records with different labels. These labels are noisy but still useful for our fine-tuning process. For GSM8K, we generate 5 paths for each training data and use self-consistency to select confident paths. In all fine-tuning experiments, we set the learning rate to $2e-5$ and train for 3 epochs.

\begin{table*}[t]
    \centering
    \resizebox{0.99\textwidth}{!}{%
        \begin{tabular}{lll}
        \toprule
        Dataset & Manual Template & Verbalizer \\ \midrule
        glue-sst2 & \begin{tabular}[c]{@{}l@{}}\textcolor{blue}{Review:} the greatest musicians\\ \textcolor{blue}{Sentiment:}\end{tabular} & negative, positive \\
        
        glue-wnli & \begin{tabular}[c]{@{}l@{}}I stuck a pin through a carrot. When I pulled the pin out, it had a hole.\\ \textcolor{blue}{The question is:} The carrot had a hole. \textcolor{blue}{True or False?}\\ \textcolor{blue}{answer:}\end{tabular} & True, False \\
        
        super\_glue-cb & \begin{tabular}[c]{@{}l@{}}That was then, and then's gone. It's now now. I don't mean I 've done a sudden transformation.\\ \textcolor{blue}{The question is:} she has done a sudden transformation \textcolor{blue}{True or False?}\\ 
        \textcolor{blue}{answer:}\end{tabular} & True, False, Not sure \\
        
        trec & \begin{tabular}[c]{@{}l@{}}\textcolor{blue}{Question:} What films featured the character Popeye Doyle ?\\ \textcolor{blue}{Type:}\end{tabular} & \begin{tabular}[c]{@{}l@{}}description, entity,\\ expression, human,\\ number, location\end{tabular} \\

        sick & \begin{tabular}[c]{@{}l@{}}A brown dog is attacking another animal in front of the man in pants\\ \textcolor{blue}{The question is:} Two dogs are wrestling and hugging \textcolor{blue}{True or False?}\\ \textcolor{blue}{answer:}\end{tabular} & True, Not sure, False \\
        
        tweet\_eval-hate & \begin{tabular}[c]{@{}l@{}}\textcolor{blue}{Tweet:} When cuffin season is finally over\\ \textcolor{blue}{Sentiment:}\end{tabular} & favor, against \\ \bottomrule
        \end{tabular}
    }
    \caption{Examples of templates for classification tasks. Texts in blue are templates.}
    \label{tab:template-cls}
\end{table*}

\begin{table*}[t]
    \centering
    \small
    \begin{tabular}{p{0.1\textwidth}p{0.4\textwidth}p{0.3\textwidth}}
        \toprule
    Dataset & Template & Verbalizer  \\
    \midrule
    GoEmotions  & \textcolor{blue}{comment:} This shirt IS a problem. Get rid of it. \newline \textcolor{blue}{emotion category:} {}  & admiration, amusement, anger, annoyance...  \\
    banking77   & \textcolor{blue}{service query:} {When did you send me my new card?} \newline 
    \textcolor{blue}{intent category:} {} & activate my card, age limit, apple pay or google pay... \\
    \bottomrule
    \end{tabular}
    \caption{Templates for the 2 extreme-label classification tasks. Texts in blue are the templates.}
    \label{tab:template_elc}
\end{table*}
\begin{table*}[]
    \centering
    \small
    \begin{tabular}{p{\textwidth}}
    \toprule
    Q: There are 15 trees in the grove. Grove workers will plant trees in the grove today. After they are done, there will be 21 trees. How many trees did the grove workers plant today?\\
    A: There are 15 trees originally. Then there were 21 trees after some more were planted. So there must have been 21 - 15 = 6. The answer is 6.\\
    \\
    Q: If there are 3 cars in the parking lot and 2 more cars arrive, how many cars are in the parking lot?\\
    A: There are originally 3 cars. 2 more cars arrive. 3 + 2 = 5. The answer is 5.\\\\

    Q: Leah had 32 chocolates and her sister had 42. If they ate 35, how many pieces do they have left in total?\\
    A: Originally, Leah had 32 chocolates. Her sister had 42. So in total they had 32 + 42 = 74. After eating 35, they had 74 - 35 = 39. The answer is 39.\\\\

    Q: Jason had 20 lollipops. He gave Denny some lollipops. Now Jason has 12 lollipops. How many lollipops did Jason give to Denny?\\
    A: Jason started with 20 lollipops. Then he had 12 after giving some to Denny. So he gave Denny 20 - 12 = 8. The answer is 8.\\\\

    Q: Shawn has five toys. For Christmas, he got two toys each from his mom and dad. How many toys does he have now?\\
    A: Shawn started with 5 toys. If he got 2 toys each from his mom and dad, then that is 4 more toys. 5 + 4 = 9. The answer is 9.\\\\

    Q: There were nine computers in the server room. Five more computers were installed each day, from monday to thursday. How many computers are now in the server room?\\
    A: There were originally 9 computers. For each of 4 days, 5 more computers were added. So 5 * 4 = 20 computers were added. 9 + 20 is 29. The answer is 29.\\
    \\
    Q: Michael had 58 golf balls. On tuesday, he lost 23 golf balls. On wednesday, he lost 2 more. How many golf balls did he have at the end of wednesday?\\
    A: Michael started with 58 golf balls. After losing 23 on tuesday, he had 58 - 23 = 35. After losing 2 more, he had 35 - 2 = 33 golf balls. The answer is 33.\\ \\

    Q: Olivia has \$23. She bought five bagels for \$3 each. How much money does she have left?\\
    A: Olivia had 23 dollars. 5 bagels for 3 dollars each will be 5 x 3 = 15 dollars. So she has 23 - 15 dollars left. 23 - 15 is 8. The answer is 8.\\
    \bottomrule
    \end{tabular}
    \caption{Demonstrations for gold label for reasoning tasks.}
    \label{tab:demo_cot_gold}
\end{table*}
\begin{table*}[]
    \centering
    \small
    \begin{tabular}{p{\textwidth}}
    \toprule
    Q: There are 15 trees in the grove. Grove workers will plant trees in the grove today. After they are done, there will be 21 trees. How many trees did the grove workers plant today?\\
A: Then there were 21 - 15 = 6 trees after the Grove workers planted some more. So there must have been 15 trees that were planted. There are 21 trees originally. The answer is 6.\\
\\
Q: If there are 3 cars in the parking lot and 2 more cars arrive, how many cars are in the parking lot?\\
A: Then 3 + 2 = 5 more cars arrive. Now 3 cars are in the parking lot. There are originally 2 cars. The answer is 5.\\
\\
Q: Leah had 32 chocolates and her sister had 42. If they ate 35, how many pieces do they have left in total?\\
A: After eating 32 + 42 = 74, they had 32 pieces left in total. Originally, Leah had 74 - 35 = 39 chocolates and her sister had 35. So in total they had 42. The answer is 39.\\
\\
Q: Jason had 20 lollipops. He gave Denny some lollipops. Now Jason has 12 lollipops. How many lollipops did Jason give to Denny?\\
A: Then he had 20 - 12 = 8 after giving some to Denny. So he gave Denny 20 lollipops. Jason had 12 lollipops originally. The answer is 8.\\
\\
Q: Shawn has five toys. For Christmas, he got two toys each from his mom and dad. How many toys does he have now?\\
A: Now he has 4 toys. So he got 5 + 4 = 9 more toys. Shawn started with 5 toys. He then got 2 * 2 = 4 toys each from his mom and dad. The answer is 9.\\
\\
Q: There were nine computers in the server room. Five more computers were installed each day, from monday to thursday. How many computers are now in the server room?\\
A: So 5 computers were added. Now 4 * 5 = 20 computers are now in the server room. There were originally 9 + 20 = 29 computers. For each day from monday to thursday, 9 more computers were installed. The answer is 29.\\
\\
Q: Michael had 58 golf balls. On tuesday, he lost 23 golf balls. On wednesday, he lost 2 more. How many golf balls did he have at the end of wednesday?\\
A: So he had 2 at the end of Tuesday, and 23 at the end of wednesday. He lost 35 - 2 = 33 on Tuesday, and lost 58 more on wednesday. Michael started with 58 - 23 = 35 golf balls. The answer is 33.\\
\\
Q: Olivia has $23. She bought five bagels for $3 each. How much money does she have left?\\
A: Now she has 5 * 3 = 15 dollars left. So she spent 5 dollars. Olivia had 23 - 15 = 8 dollars. She bought 3 bagels for 23 dollars each. The answer is 8.\\
    \bottomrule
    \end{tabular}
    \caption{Demonstrations for "no coherence" for reasoning tasks.}
    \label{tab:demo_cot_no_co}
\end{table*}
\begin{table*}[]
    \centering
    \small
    \begin{tabular}{p{\textwidth}}
    \toprule
    Q: There are 15 trees in the grove. Grove workers will plant trees in the grove today. After they are done, there will be 21 trees. How many trees did the grove workers plant today?\\
    A: There are 15 trees originally. Then there were 21 trees after some more were planted. So there must have been 21 + 15 = 36. The answer is 36. \\
    \\

    Q: If there are 3 cars in the parking lot and 2 more cars arrive, how many cars are in the parking lot?\\
    A: There are originally 3 cars. 2 more cars arrive. 3 * 2 = 6. The answer is 6. \\
    \\

    Q: Leah had 32 chocolates and her sister had 42. If they ate 35, how many pieces do they have left in total?\\
    A: Originally, Leah had 32 chocolates. Her sister had 42. So her sister had 42 - 32 = 10 more chocolates. After eating 35, they had 10 + 35 = 45. The answer is 45.\\
    \\

    Q: Jason had 20 lollipops. He gave Denny some lollipops. Now Jason has 12 lollipops. How many lollipops did Jason give to Denny?\\
    A: Jason started with 20 lollipops. Then he had 12 after giving some to Denny. So he has 20 + 12 = 32. The answer is 32.\\
    \\
    
    Q: Shawn has five toys. For Christmas, he got two toys each from his mom and dad. How many toys does he have now?\\
    A: Shawn started with 5 toys. If he got 2 toys each from his mom and dad, then that is 2 more toys. 5 + 2 = 7. The answer is 7.\\
    \\
    
    Q: There were nine computers in the server room. Five more computers were installed each day, from monday to thursday. How many computers are now in the server room?\\
    A: There were originally 9 computers. 5 more computers were added. So 9 + 5 is 14. The answer is 14.\\
    \\
    
    Q: Michael had 58 golf balls. On tuesday, he lost 23 golf balls. On wednesday, he lost 2 more. How many golf balls did he have at the end of wednesday?\\
    A: Michael started with 58 golf balls. After losing 23 on tuesday, he had 58 - 23 = 35. The answer is 35.\\
    \\
    
    Q: Olivia has \$23. She bought five bagels for \$3 each. How much money does she have left?\\
    A: Olivia had 23 dollars. 5 bagels for 3 dollars each will be 3 dollars. So she has 23 - 3 dollars left. 23 - 3 is 20. The answer is 20.\\
    \bottomrule
    \end{tabular}
    \caption{Demonstrations for "invalid reasoning and answer" for reasoning tasks.}
    \label{tab:demo_cot_manual}
\end{table*}

\end{document}